\documentclass[11pt]{article}
\usepackage{acl2016}
\usepackage{times}
\usepackage{url}
\usepackage{latexsym}
\usepackage{CJKutf8}
\usepackage{booktabs}
\usepackage{amsmath}
\usepackage{amssymb}
\usepackage{enumitem}
\usepackage{avisil}
\usepackage{graphicx}
\usepackage{multirow}
\usepackage{xcolor}

\aclfinalcopy 

\title{One for All: Towards Language Independent Named Entity Linking}

\author{Avirup Sil \and Radu Florian\\
  IBM T. J. Watson Research Center \\
  1101 Kitchawan Road \\
  Yorktown Heights, NY 10598 \\
  {\tt avi@us.ibm.com, raduf@us.ibm.com} }

\date{}

\newcommand{\liel}{{\sc LieL}}
\renewcommand{\etal}{\textit{et al.}}
\newcommand{\jnerd}{{\sc Nerel}}

\newcommand{\feature}[1]{\textbf{\textsc{#1}}}

\begin{document}
\begin{CJK}{UTF8}{gbsn}%
\end{CJK} \inputencoding{latin9}
\maketitle
\begin{abstract}
Entity linking (EL) is the task of disambiguating mentions in text
by associating them with entries in a predefined database of mentions
(persons, organizations, etc). Most previous EL research
has focused mainly on one language, English, with less attention being
paid to other languages, such as Spanish or Chinese. In this paper, we introduce \liel, a Language Independent
Entity Linking system, which provides an EL framework which, once
trained on one language, works remarkably well
on a number of different languages without change. \liel\ makes
a joint global prediction over the entire document, employing a discriminative
re-ranking framework with many domain and language-independent
feature functions. Experiments on numerous benchmark datasets,
show that the proposed system, once trained on one language, English,
outperforms several state-of-the-art systems in English (by 4 points)
and the trained model also works very well on Spanish (14 points better than a competitor system), demonstrating the viability of the approach.

\end{abstract}
\section{Introduction}

We live in a golden age of information, where we have access to vast
amount of data in various forms: text, video and audio. Being able to analyze
this data automatically, usually involves filling a relational database,
which, in turn, requires the processing system to be able
to identify actors across documents by assigning unique identifiers to them. 
Entity Linking (EL) is the task of mapping specific textual mentions of entities in
a text document to an entry in a large catalog of entities, often
called a knowledge base or KB, and is one of the major tasks in the
Knowledge-Base Population track at the Text Analysis Conference (TAC) \cite{ji2014overview}.
The task also involves grouping together (clustering) $NIL$ entities
which do not have any target referents in the KB.

Previous work, pioneered by \cite{2006-eacl-bunescu-wikipedia-ne-disambig,2007-emnlp-conll-cucerzan-wikipedia-NE-disambig,2012-emnlp-open-ne-resolution,2011-acl-illinois-wikifier,2013-naacl-twitter-nerd},
have used Wikipedia as this target catalog of entities because of
its wide coverage and its frequent updates made by the community. As
with many NLP approaches, most of the previous EL research
have focused on English, mainly because it has many NLP resources
available, it is the most prevalent language on the web, and the fact
that the English Wikipedia is the largest among all the Wikipedia datasets.
However, there are plenty of web documents in other languages, such
as Spanish \cite{fahrni2013hits,ji2014overview}, and Chinese \cite{rpi-emnlp14,rpi-acl14},
with a large number of speakers, and there is a need to be able to
develop EL systems for these languages (and others!) quickly and inexpensively.

In this paper, we investigate the hypothesis that we can train an EL model that is entirely un-lexicalized, by only allowing features that compute similarity between the text in the input document and the text/information in the KB. For this purpose, we propose a novel approach to entity linking, which we call Language Independent Entity Linking (henceforth
\liel). 
We test this hypothesis by applying the English-trained system on Spanish and Chinese datasets, with great success. 


This paper has three novel contributions: 1) extending a powerful inference algorithm
for global entity linking, built using similarity measures, 
corpus statistics, along with knowledge base statistics, 2) integrates many
language-agnostic and domain independent features in an
exponential framework, and 3) provide empirical evidence on a large variety of 
popular benchmark datasets that the resulting model outperforms or matches the best published results, and, most importantly, the trained model transfers well across languages, outperforming the state-of-the-art (SOTA) in Spanish and matching it in Chinese.

We organize the paper as follows: the next section motivates the problem and discusses the language-independent model along with the features. Section \ref{sec:experiments} describes our experiments and comparison with the state-of-the-art. Section \ref{sec:related-work} illustrates the related previous work and Section \ref{sec:conclusion} concludes.

\section{Problem Formulation\label{sec:problem-formulation}}

\subsection{Motivation for Language Independence}

Our strategy builds an un-lexicalized EL system
by training it on labeled data, which consists of pairs of mentions in text and entries
in a database extracted from a Wikipedia collection in English.
Unlike traditional EL, however, the purpose here is
to be able to perform entity linking with respect to any Wikipedia collection.
Thus the strategy must take care to build a model that can transfer its learned model
to a new Wikipedia collection, without change.

At a first glance, the problem seems very challenging - learning how to discriminate
\texttt{Lincoln,\_Nebraska} and \texttt{Abraham\_Lincoln} \footnote{Teletype font denotes Wikipedia titles and categories.},
the former US President, seemingly bears little resemblance to
disambiguating between different Spanish person entities named ``Ali Quimico''. The 
crux of the problem lies in the fact that Wikipedia-driven
features are language-specific: for instance, counting how many times the category
\texttt{2010 Deaths} appears in the context of an entity is highly
useful in the English EL task, but not directly useful for Spanish EL.
Also, learning vocabulary-specific information like the list of ``deaths'',
``presidents'', etc. is very useful for disambiguating person entities
like ``Lincoln'' in English, but the same model, most likely, will
not work for mentions like ``\begin{CJK*}{UTF8}{gbsn}李娜\end{CJK*}''  in a Chinese document
which might either refer to the famous athlete \begin{CJK*}{UTF8}{gbsn}李娜\ (网球运动员)\end{CJK*} or the singer
\begin{CJK*}{UTF8}{gbsn}李娜\ (歌手)\end{CJK*}.

Practically we assume the existence of a knowledge
base that defines the space of entities we want to disambiguate against, where
each entry contains a document with the entity; Wikipedia
is a standard example for this\footnote{We will assume, without loss of generality, that the knowledge base
is derived from Wikipedia.}. If there are other properties associated
with the entries, such as categories, in-links, out-links, redirects,
etc., the system can make use of them, but they are theoretically
not required. The task is defined as: given a mention $m$ in a document
$d$, find the entry $e$ in the knowledge base that $m$ maps to.

We expand on the architecture described in \cite{cikm-joint-nerel} (henceforth \jnerd), because of the flexibility provided by the feature-based exponential framework which results in an English SOTA EL system. 
However, we design all our features in such a way that they measure
the similarity between the context where the mention $m$ appears in $d$ and the
entries in the knowledge base. For example, instead of
counting how often the category \texttt{2010 Deaths} \footnote{Or a specific Freebase type.} appears in the
context around an entity mention, we create a feature function such
as \feature{Category Frequency}$(m, e)$, which counts how often any
category of entity referent $e$ appears in the context of mention
$m$. For entities like \texttt{Lincoln,\_Nebraska} in the English
EL, \feature{Category Frequency} will add together counts for appearances
of categories like \texttt{Cities in Lancaster County, Nebraska} and
\texttt{Lincoln metropolitan area}, among other categories. At the
same time, in the Spanish EL domain, \feature{Category Frequency}
will add together counts for \texttt{Pol\'{i}ticos de Irak} and \texttt{Militares
de Irak} for the KB id corresponding to ``Ali Quimico''. This feature
is well-defined in both domains, and larger values of the feature
indicate a better match between $m$ and $e$. As mentioned earlier, it is our hypothesis,
that the parameters trained for such features on one language (English, in our case) can be successfully
used, without retraining, on other languages, namely Spanish and Chinese.

While training, the system will take as input a knowledge base in source language $S$, $KB_{S}$ (extracted from Wikipedia)
and a set of training examples $(m_{i},e_{i},g_{i})$, where instances
$m_{i}$ are mentions in a document of language $S$, $e_{i}$ are entity
links, $e_{i}\in KB_{S}$, and $g_{i}$ are Boolean values indicating the gold-standard match / mismatch between $m_{i}$ and $e_{i}$. During decoding, given language $T$\footnote{Language prediction can be done relatively accurately, given a document; however, in this paper, we focus on the EL task, so we assume we know the identity of the target language $T$.},
the system must classify examples $(m_{j},e_{j})$ drawn from a \emph{target}
language $T$ and knowledge-base $KB_{T}$.

\subsection{\liel: Training and Inference}

Our language-independent system consists of two components: 1. extracting
mentions of named-entities from documents and 2. linking the detected
mentions to a knowledge base, which in our case is Wikipedia (focus of this paper). We run the IBM Statistical Information and Relation Extraction (SIRE) \footnote{The IBM SIRE system can be currently accessed at :\\http://www.ibm.com/smarterplanet/us/en/ibmwatson/\\developercloud/relationship-extraction.html} system which is a toolkit that performs mention detection, relation extraction, coreference resolution, \etc\  We use the system to extract mentions and perform
coreference resolution: in particular, we use the CRF model of IBM SIRE 
for mention detection and a maximum entropy clustering
algorithm for coreference resolution. The system identifies a set
of 53 entity types. To
improve the mention detection and resolution, case restoration is
performed on the input data. Case restoration is helpful to improve the mention detection system's performance, especially for discussion forum data.
Obviously, this processing step is language-dependent, as the information extraction system is - but we want to emphasize that the entity linking system is language independent.

In the EL step, we perform a full document entity disambiguation
inference, described as follows. Given a document $d$, and a selected mention $m\in d$, our goal is to identify its label $\hat{e}$ that maximizes{\footnotesize{}
\begin{eqnarray}
\hat{e} & = & P\left(e|m,d\right)\label{eq:globalinference}\\
 & = & \arg\max_{e:m}\sum_{k,m\in m_{1}^{k},e_{1}^{k}}P\left(m_{1}^{k}|m,d\right)P\left(e_{1}^{k}|m_{1}^{k},d\right)\nonumber 
\end{eqnarray}
}where $m_{1}^{k}$ are mentions found in document $d$, and $e_{1}^{k}$
are some label assignment. In effect, we are looking for the best mention labeling of the entire document $m_{1}^k$ (that contains $m$) and a label to these mentions that would maximize the information extracted from
the entire document. Since direct inference on Equation \ref{eq:globalinference}
is hard, if not intractable, we are going to select the most likely
mention assignment instead (as found by an information extraction
system): we will only consider the detected mentions $\left(m_{1},\ldots,m_{k}\right)$,
and other optional information that can be extracted from the document, such
as links $l$, categories $r$, etc.
The goal becomes identifying the set of labels $\left(e_{1},\ldots,e_{k}\right)$
that maximize  
\begin{equation}
P\left(e_{1}^{k}|m_{1}^{k},d\right)\label{eq:entity-linking}
\end{equation} 
Since searching over all possible sets of (mention, entity)-pairs
for a document is still intractable for reasonable large values
of $k$, typical approaches to EL make simplifying assumption on how
to compute the probability in Equation \ref{eq:entity-linking}. Several
full-document EL approaches have investigated generating up to $N$
global tuples of entity ids $(e_{1},\ldots,e_{k})$, and then build
a model to rank these tuples of entity ids \cite{2006-eacl-bunescu-wikipedia-ne-disambig,2007-emnlp-conll-cucerzan-wikipedia-NE-disambig}.
However, Ratinov \etal\ \cite{2011-acl-illinois-wikifier} argue
that this type of global model provides a relatively small improvement
over the purely-local approach (where $P\left(e_{1}^{k}|m_{1}^{k},d\right)=\prod_{i}P\left(e_{i}|m_{i},d\right)$).
In this paper, we follow an approach which combines both of these
strategies. 

Following the recent success of \cite{cikm-joint-nerel},
we partition the full set of extracted mentions, $\left(m_{i}\right)_{i=\bar{1,n}}$
of the input document $d$ into smaller subsets of mentions which
appear near one another: we consider two mentions that are closer
then 4 words to be in the same connected component, then we take the
transitive closure of this relation to partition the mention set.
We refer to these sets as the \emph{connected components} of $d$,
or $CC(d)$. We perform classification over the set of entity-mention
tuples $T\left(C\right)=\left\{ \left(e_{i_{1}},\ldots,e_{i_{n_{C}}}|m_{i_{1}},\ldots,m_{i_{n_{C}}}\right)|e_{i_{j}}\in KB,\forall j\right\} $
\footnote{For simplicity, we denote by $(e|m)$ the tuple $(e,m)$, written like that to capture the fact that $m$ is fixed, while $e$ is predicted.}
that are formed using candidate entities within the same connected
component $C\in CC(d)$. Consider this small snippet of text:  \vspace{-2mm}
\begin{quote}
\hspace{-5mm} ``\ldots{}Home Depot CEO Nardelli quits \ldots{}'' 
\end{quote}
In this example text, the phrase ``Home Depot CEO Nardelli'' would
constitute a connected component. Two of
the entity-mention tuples for this connected component would be: (\texttt{Home\_Depot}, \texttt{Robert\_Nardelli} $|$''Home Depot'',
``Nardelli'') and (\texttt{Home\_Depot}, \texttt{Steve\_Nardelli} $|$ ''Home Depot'',``Nardelli'').

\subsubsection{Collective Classification Model}
To estimate $P(t|d,C)$, the probability
of an entity-mention tuple $t$ for a given connected component $C\in CC(d)$, \liel\
uses a maximum-entropy model:
\begin{equation}
P(t|d,C)=\frac{\exp\left(\vect{w}\cdot\vect{f}(t,d,C)\right)}{\sum_{t'\in T(C)}\exp\left(\vect{w}\cdot\vect{f}(t',d,C)\right)}
\label{eq:tuple-prob}
\end{equation}
where $\vect{f}(t,d,C)$ is a feature vector associated with $t$, $d$, and $C$, and $w$ is a weight vector.
For training, we use L2-regularized conditional log likelihood (CLL) as the objective
\begin{equation}
CLL(G,\vect{w})=\sum_{(t,d,C)\in G}\log P(t|d,C,\vect{w})-\sigma\|\vect{w}\|_{2}^{2}\label{CLL}
\end{equation}
where $G$ is the gold-standard training data, consisting of pairs
$(t,d,C)$, where $t$ is the correct tuple of entities and mentions
for connected component $C$ in document $d$, and $\sigma$ is a
regularization parameter. Given that the function \ref{CLL} is convex,
we use LBFGS \cite{LBFGS} to find the globally optimal parameter
settings over the training data.

\subsection{Extracting potential target entities}
\label{sec:fast-match}
From the dump of our Wikipedia data, we extract all the mentions that 
can refer to Wikipedia titles, and construct a set
of disambiguation candidates for each mention (which are basically the hyperlinks
in Wikipedia). 
This is, hence, an anchor-title index that
maps each distinct hyperlink anchor-text to its corresponding
Wikipedia titles and also stores their relative popularity score. For example, the anchor text (or mention)
``Titanic" is used in Wikipedia to refer both to the
ship or to the movie.
To retrieve the disambiguation candidates $e_i$ for
a given mention $m_i$, we
query the anchor-title index that we constructed and use
lexical sub-word matching. $e_i$
is taken to be the
set of titles (or entities, in the case of EL) most frequently linked to with anchor text $m_i$
in Wikipedia. We use only the top 40 most frequent Wikipedia candidates
for the anchor text for computational efficiency purposes for most of our
experiments. We call this step ``Fast Search'' since it produces a bunch of candidate links
by just looking up an index.

\subsubsection{Decoding}
At decoding time, given a document $d$, we identify its connected
components $CC\left(d\right)$ and run inference on each component
$C$ containing the desired input mention $m$. To further reduce
the run time, for each mention $m_{j}\in C$, we obtain the set of potential 
labels $e_{j}$ using the algorithm described in Section \ref{sec:fast-match},
and then exhaustively find the pair that maximizes equation \ref{eq:tuple-prob}. For each candidate link,
we also add a NIL candidate to fast match to let the system link mentions to ids not in a KB.

\subsection{Language-Independent Feature Functions}

\liel\ 
makes use of new as well as well-established features in the EL literature. However, we make sure to use only non-lexical features. The local and global feature functions computed from this extracted information are described below.

Generically, we have two types of basic features: one that takes as input a KB entry $e$, the mention $m$ and its document and a second type that scores two KB entries, $e_1$ and $e_2$. When computing the probability in Equation \ref{eq:tuple-prob}, where we consider a set of KB entries $t$\footnote{Recall that the probability is computed for all the entity assignments for mentions in a clique.}, we either sum or apply a boolean \cal{AND} operator (in case of boolean features) among all entities $e\in t$, while the entity-entity functions are summed/and'ed for consecutive entities in $t$. We describe the features in these terms, for simplicity. 

\subsubsection{Mention-Entity Pair Features}

\noindent\textbf{Text-based Features:} We assume the existence of a document with most entries in the KB, and the system uses similarity between the input document and these KB documents.  The basic intuition behind these features, inspired by Ratinov \etal \shortcite{2011-acl-illinois-wikifier},
is that a mention $m\in d$ is more likely to refer to entity $e$
if its KB page, $W(e)$, has high textual similarity to input document $d$. 
Let $Text\left(W\left(e\right)\right)$ be the vector
space model associated with $W\left(e\right)$, $Top\left(W\left(e\right)\right)$
be the vector of the top most frequently occurring words (excluding
stop-words) from $W\left(e\right)$, and $Context(W(e))$ be the vector space of the 100 word window around the first occurrence of $m$ in $W(e)$.
Similarly, we create vector space models $Text(m)$ and $Context(m)$.
We then use cosine similarity over these vector space models as features:\\
i.  cosine$\left(Text\left(W\left(e\right)\right),Text\left(m\right)\right)$,\\
ii. cosine$\left(Text\left(W\left(e\right)\right),Context\left(m\right)\right)$,\\
iii. cosine$\left(Context\left(W\left(e\right)\right),Text\left(m\right)\right)$,\\
iv. cosine$\left(Context\left(W\left(e\right)\right),Context\left(m\right)\right)$,\\
v. cosine$\left(Top\left(W\left(e\right)\right),Text\left(m\right)\right)$,\\
vi. cosine $\left(Top\left(W\left(e\right)\right),Context\left(m\right)\right)$.\\

\noindent\textbf{KB Link Properties:} \liel\ can make use of existing relations in the KB, such as inlinks, outlinks, redirects, and categories. Practically, for each such relation $l$, a KB entry $e$ has an associated set of strings $I(l,e)$\footnote{For instance, redirect strings for ``Obama" are ``Barack Obama", ``Barack Obama Jr." and ``Barack Hussein Obama".}; given a mention-side set $M$ (either $Text(m)$ or $Context(m)$), \liel\
computes \feature{Frequency}
feature functions for the names of the Categories, Inlinks, Outlinks and Redirects, we compute 
\[f(e, m, d) = \left|I(l,e) \cap M\right|\]

\noindent\textbf{Title Features:} \liel\ also contains a number of features
that make use of the Wikipedia title of the entity links in $t$ (remember $t$ = entity mention
tuples and not a Wikipedia title) :

\begin{itemize}[noitemsep,nolistsep]
\item \feature{Nil Frequency}:  Computes the frequency of entities that link
to $NIL$
\item \feature{Exact Match Frequency}: returns 1 if the surface form of $m$ is a redirect for $e$; 
\item \feature{Match\ All}: returns true if $m$ matches \textit{exactly} the title of $e$;
\item \feature{Match Acronym}: returns true if $m$ is an acronym for a redirect of $e$;
\item  \feature{Link Prior}: the prior link probability $P(e|m)$, computed from anchor-title pairs in KB (described in Section \ref{sec:fast-match}).
\end{itemize}

\subsubsection{Entity-Entity Pair Features}

\noindent\textbf{Coherence Features:} To better model consecutive entity assignments, \liel\ computes a coherence feature function called \feature{Outlink Overlap}. 
For every consecutive pair of entities $(e_1,e_2)$
that belongs to mentions in $t$, the feature 
computes ${\displaystyle \operatorname{Jaccard}(Out(e_1), Out(e_2))}$,
where $Out(e)$ denotes the Outlinks of $e$. Similarly, we also compute \feature{Inlink Overlap}.

\liel\ also uses categories in Wikipedia which exist in all languages.
The first feature \feature{Entity Category PMI}, inspired by Sil and Yates \shortcite{cikm-joint-nerel}, 
make use of Wikipedia's
category information system to find patterns of entities that commonly
appear next to one another. Let $\mathcal{C}(e)$ be the set of Wikipedia categories
for entity $e$. We manually inspect and remove a handful of common Wikipedia categories based on threshold
frequency on our training data, which are associated
with almost every entity in text, like \texttt{Living People} etc.,
since they have lower discriminating power. These are analogous to all WP languages. From the training data,
the system first computes point-wise mutual information (PMI) \cite{turney2002}
scores for the Wikipedia categories of pairs of entities, $(e_{1},e_{2})$:
\begin{multline*}
PMI(\mathcal{C}(e_{1}),\mathcal{C}(e_{2}))=\\
\frac{{\displaystyle \sum_{j=1}^{n_{t_{\mathcal{C}}}-1}\mathbf{1}[\mathcal{C}(e_{1})=\mathcal{C}(e_{i_{j}})\wedge \mathcal{C}(e_{2})=\mathcal{C}(e_{i_{j+1}})]}}{{\displaystyle \sum_{j}\mathbf{1}[\mathcal{C}(e_{1})=\mathcal{C}(e_{i_{j}})]\times\sum_{j}\mathbf{1}[\mathcal{C}(e_{2})=\mathcal{C}(e_{i_{j}})]}}
\end{multline*}

\begin{itemize}[noitemsep,nolistsep]
\item \feature{Entity Category PMI} adds these PMI scores up for every consecutive
$(e_{1},e_{2})$ pair in $t$. 
\item \feature{Categorical Relation Frequency} We would like to boost consecutive entity assignments that have been seen in the training data. For instance, for the text ``England captain Broad fined for..'',
we wish to encourage the tuple that links ``England'' to the entity
id of the team name \texttt{England cricket team}, and ``Broad''
to the entity id of the person \texttt{Stuart Broad}. Wikipedia contains
a relation displayed by the category called \texttt{English\_cricketers}
that indicates that \texttt{Stuart Broad} is a team member of \texttt{England
cricket team}, and counts the number of such relations between every
consecutive pair of entities in $(e,e')\in t$. 
\item \feature{Title Co-occurrence Frequency} feature computes for
every pair of consecutive entities $(e,e')\in t$, the number of times
that $e'$ appears as a link in the Wikipedia page for $e$, and \emph{vice
versa} (similar to \cite{2007-emnlp-conll-cucerzan-wikipedia-NE-disambig}. It adds these counts up to get a single number for $t$.\end{itemize}

\section{Experiments}

\label{sec:experiments} 
We evaluate \liel's capability by testing against several state-of-the-art EL systems on English, then apply the English-trained system to Spanish and Chinese EL tasks to test its language transcendability.

\subsection{Datasets\label{sub:Datasets}}

\textbf{English:} The 3 benchmark datasets for
the English EL task are: \textbf{i)} ACE \cite{2011-acl-illinois-wikifier},
\textbf{ii)} MSNBC \cite{2007-emnlp-conll-cucerzan-wikipedia-NE-disambig}
and \textbf{iii)} TAC 2014 \cite{ji2014overview}\footnote{This is the traditional Entity Linking (EL) task and not Entity Discovery and Linking (EDL), since we are comparing the linking capability in this paper.}, which contain data
from diverse genre like discussion forum, blogs and news. 
Table \ref{table:labeled-data-stats} provides
key statistics on these datasets. 
In the TAC\footnote{For more details on TAC see http://nlp.cs.rpi.edu/kbp/2014/index.html}
evaluation setting, EL systems are given as input a document and a
query mention with its offsets in the input document. As the output,
systems need to predict the KB id of the input query mention if it
exists in the KB or $NIL$ if it does not. Further, they need to cluster
the mentions which contain the same $NIL$ ids across queries.

The training dataset, WikiTrain, consists of 10,000 random Wikipedia
pages, where all of the phrases that link to other Wikipedia articles
are treated as mentions, and the target Wikipedia page is the label.
The dataset was made available by Ratinov \etal\ and \cite{cikm-joint-nerel}, 
added Freebase to Wikipedia mappings resulting in
158,715 labeled mentions with an average of 12.62 candidates
per mention. The total number of unique mentions in the data set is
77,230 with a total of 974,381 candidate entities and 643,810 unique
candidate entities. The Wikipedia dump that we used as our knowledge-base
for English, Spanish and Chinese is the April 2014 dump. 
The TAC dataset involves the TAC KB which is a dump of May 2008
of English Wikipedia. \liel\ links entities to the Wikipedia 2014
dump and uses the redirect information to link back to the TAC KB.

\begin{table}
\begin{centering}
\begin{tabular}{lccc}
\toprule 
Name & $|M|$  & In\ KB  & Not\ in\ KB \tabularnewline
\midrule 
ACE  & 257  & 100\%  & 0 \tabularnewline
MSNBC  & 747  & 90\%  & 10\% \tabularnewline
TAC\_En14  & 5234  & 54\%  & 46\% \tabularnewline
TAC\_Es13  & 2117  & 62\%  & 38\% \tabularnewline
TAC\_Es14  & 2057  & 72\%  & 28\% \tabularnewline
TAC\_Zh13  & 2155  & 57\%  & 43\% \tabularnewline
WikiTrain  & 158715  & 100\%  & 0\%\tabularnewline
\bottomrule
\end{tabular}
\par\end{centering}

\protect\caption{Data statistics: number of mention queries, \% of mention queries that have their referents
present in the Wikipedia/KB, and \% of mention queries that have no
referents in Wikipedia/KB as per our datasets. En=English, Es=Spanish and Zh=Chinese for the evaluation
data for TAC for the years 2013 and 2014.\label{table:labeled-data-stats} }
\end{table}

\textbf{Spanish:} We evaluate \liel\ on
both the 2013 and 2014 benchmark datasets of the TAC Spanish evaluation.

\textbf{Chinese:} We  test \liel\ on the TAC 2013 Chinese dataset. 

 \vspace{-1mm}
\subsection{Evaluation Metric}

We follow standard measures used in the literature for the entity
linking task. To evaluate EL accuracy on ACE and MSNBC, we report
on a Bag-of-Titles (BOT) F1 evaluation as introduced by \cite{2008-cikm-milne-witten-wikifier,2011-acl-illinois-wikifier}.
In BOT-F1, we compare the set of Wikipedia titles output for a document
with the gold set of titles for that document (ignoring duplicates),
and compute standard precision, recall, and F1 measures. On the TAC
dataset, we use standard metrics $B^{3}+$ variant of precision, recall
and F1. On these datasets, the $B^{3}+F1$ metric includes the clustering
score for the $NIL$ entities, and hence systems that only perform
binary $NIL$ prediction would be heavily penalized\footnote{For more details on the scoring metric used for TAC EL see: http://nlp.cs.rpi.edu/kbp/2014/scoring.html}.

\subsection{Comparison with the State-of-the-art}

To follow the guidelines for the TAC NIST evaluation, we anonymize participant system names as System 1 through 9. Interested readers may look at their system description and scores in \cite{ji2014overview,fahrni2013hits,fujitsu2013tac,mayfield2013tac,merhav2013basis}. Out of these systems, System 1 and System 7 obtained the top score in Spanish and Chinese EL evaluation at TAC 2013 and hence can be treated as the current state-of-the-art for the respective EL tasks. We also compare \liel\ with some traditional ``wikifiers" like MW08 \cite{2008-cikm-milne-witten-wikifier} and UIUC \cite{emnlp-chengroth-2013} and also \jnerd\ \cite{cikm-joint-nerel} which is the system which \liel\ resembles the most.




\subsection{Parameter Settings}

\liel\ has two tuning parameters: $\sigma$, the regularization
weight; and the number of candidate links per mention we select from the Wikipedia
dump. We set the value of $\sigma$ by trying five possible values
in the range {[}0.1, 10{]} on held-out data (the TAC 2009 data). We
found $\sigma=0.5$ to work best for our experiments. We chose to
select a maximum of 40 candidate entities from Wikipedia for each
candidate mention (or fewer if the dump had fewer than 40 links with
nonzero probability).

\subsection{Results}


\begin{figure}[t]
\begin{center}
\fbox{\includegraphics[trim = 5mm 3mm 5mm 5mm, clip, width=0.98\columnwidth]{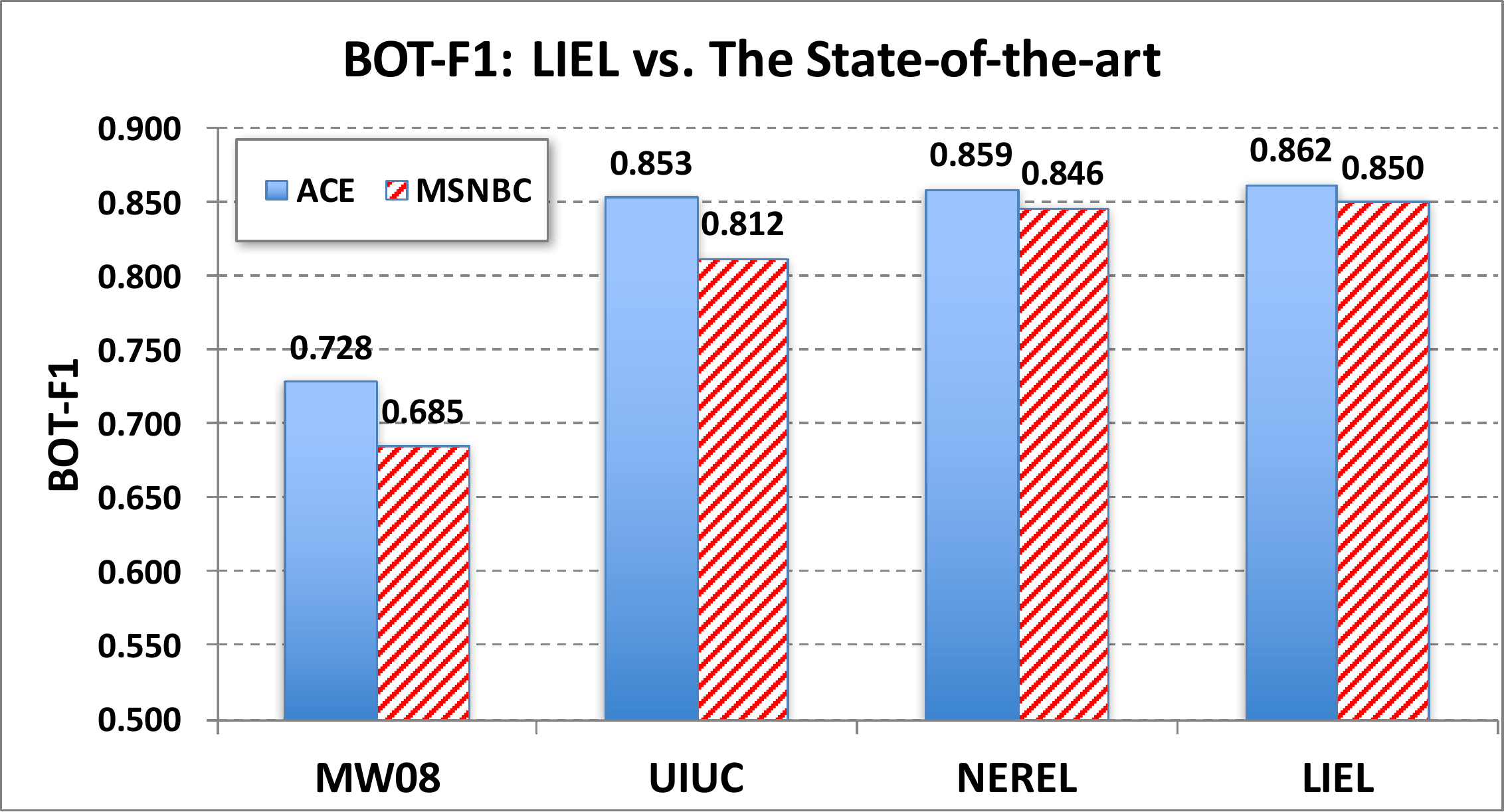}}
\end{center}
\vspace{-3mm}
\caption{\liel\ outperforms all its competitors on both ACE and MSNBC.}
\label{fig:el-results-english}
\vspace{-3mm}
\end{figure}

%

\textbf{English:} Figure \ref{fig:el-results-english} compares \liel\ with previously
reported results by MW08, UIUC  and \jnerd\ on the ACE and MSNBC
datasets in \cite{emnlp-chengroth-2013,cikm-joint-nerel}. \liel\ achieves an F1 score of 86.2 on ACE and 85.0 on
MSNBC, clearly outperforming the others \eg\ 3.8\% absolute value higher than UIUC on MSNBC. We believe that \liel's strong model comprising relational information (coherence features from large corpus statistics), textual and title lets it outperform UIUC and MW08 where the former uses relational information and the latter a naive version of \liel's coherence features. Comparison with \jnerd\ is slightly unfair (though we outperform them marginally) since they use both Freebase and Wikipedia as their KB whereas we are comparing with systems which only use Wikipedia as their KB.

\begin{figure}[t]
\begin{center}
\fbox{\includegraphics[scale=0.25, trim = 10mm 23mm 22mm 25mm, clip]{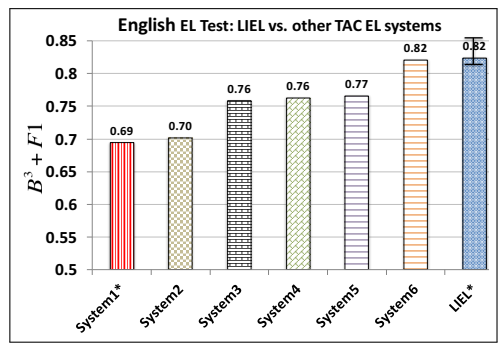}}
\end{center}
\vspace{-3mm}
\caption{Comparison of several state-of-the-art English EL systems along with
\liel\ on the latest TAC 2014 dataset and \liel\ obtains the best
score. {*} indicates systems that perform multilingual EL.}
\label{fig:el-results-english-tac}
\vspace{-3mm}
\end{figure}

To test the robustness of \liel\ on a diverse genre of data, we
also compare it with some of the other state-of-the-art
systems on the latest benchmark TAC 2014 dataset. Figure \ref{fig:el-results-english-tac} shows our results when compared with the top systems in the evaluation. Encouragingly, \liel's performance is tied with the top performer, System 6, and outperforms all the other top participants from this challenging annual evaluation. Note that \liel\ obtains 0.13 points more than System 1, the only other multi-lingual EL system and, in that sense, \liel's major competitor. Several other factors are evident from the results: System 1 and 2 are statistically tied and so are System 3, 4 and 5. We also show the bootstrapped percentile confidence intervals \cite{singh2008bootstrap} for \liel\ which are [0.813, 0.841]: (we do not have access to the other competing systems). 

\subsubsection{Foreign Language Experiments}
Note that \liel\ was \emph{trained only} on the English Wikitrain dataset (Section \ref{sub:Datasets}),
and then applied, \emph{unchanged}, to all the evaluation datasets across languages and
domains described in Section \ref{sub:Datasets}. Hence, it is the same instance of the model for all languages. As we will observe, this one system consistently outperforms the state of the art, even
though it is using exactly the same trained model across the datasets.
We consider this to be the take-away message of this paper.

\begin{figure}[t]
\begin{center}
\fbox{\includegraphics[scale=0.25, trim = 10mm 12mm 22mm 15mm, clip]{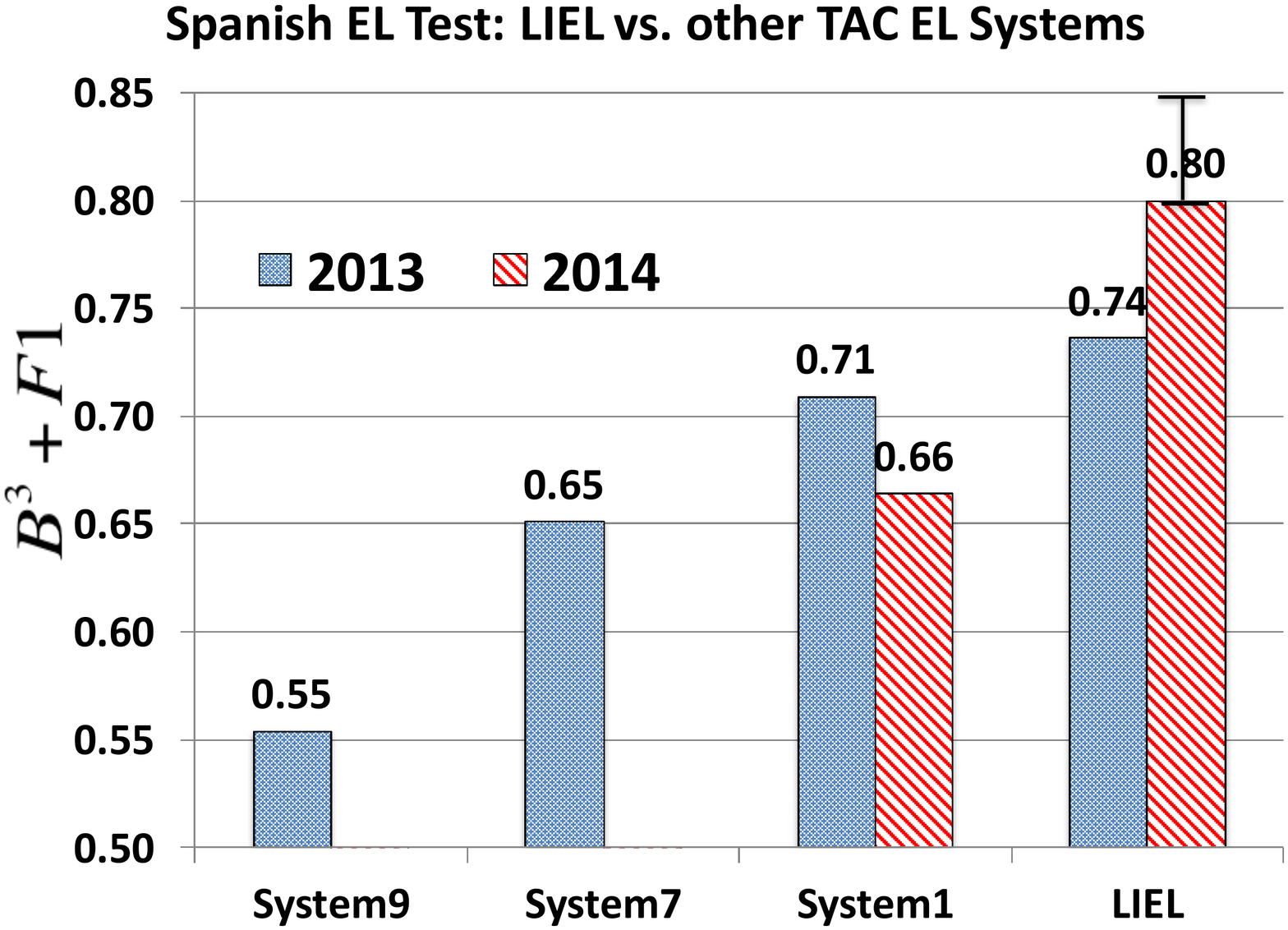}}
\end{center}
\vspace{-4mm}
\caption{System performance on the TAC 2013 and 2014 Spanish datasets are shown. 
\liel\ outperforms all the systems in terms of overall F1 score. }
\label{fig:el-results-spanish-2013}
\vspace{-3mm}
\end{figure}

\noindent\textbf{Spanish: }\liel\ obtains a $B^{3}+F1$ score of 0.736 on the TAC 2013 dataset
and clearly outperforms the SOTA, System 1, which obtains 0.709 as shown in Figure \ref{fig:el-results-spanish-2013}
and considerably higher than the other participating systems. We could only obtain the results for Systems 9 and 7 on 2013. On the 2014 evaluation dataset, \liel\ obtains a higher gain of 0.136 points (precision of 0.814 and recall of 0.787) over its major competitor System 1, showing the power of its language-independent model. 


\begin{figure}[t]
\begin{center}
\fbox{\includegraphics[scale=0.25, trim = 12mm 15mm 25mm 20mm, clip]{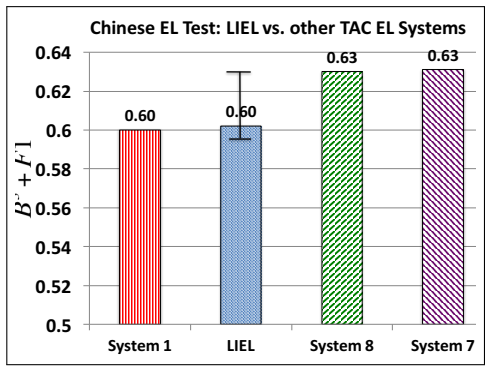}}
\end{center}
\vspace{-4mm}
\caption{\liel\ achieves competitive performance in Chinese EL further
proving its robustness to multilingual data.}
\label{fig:el-results-chinese-2013}
\vspace{-3mm}
\end{figure}

\noindent\textbf{Chinese:} Figure \ref{fig:el-results-chinese-2013} shows the results of \liel's
performance on the Chinese benchmark dataset compared to the state-of-the-art. Systems 7 and 8 obtains almost similar scores. We observe that \liel\ is tied with System 1 and achieves competitive performance compared to Systems 7 and 8 (note that \liel\ has a confidence interval of [0.597, 0.632]) which requires labeled Chinese TAC data to be trained on and the same model does not work for other languages. Emphasizing again: \liel\ is trained only \emph{once}, on English, and tested on Chinese unchanged.

\subsubsection{Error Analysis}

While we see \liel's strong multi-lingual empirical results, it is important to note some of the areas which confuses the system. Firstly, a major source of error which affects \liel's performance is due to coreference resolution \eg\ from the text ``\emph{Beltran Leyva, also known as ``The Bearded One,'' is ...}'', TAC's mention query asks the systems to provide the disambiguation for \emph{The Bearded One}. \liel\ predicts that the \emph{The Bearded One} refers to the entity \texttt{Richard Branson}, which is the most common entity in Wikipedia that refers to that nickname (based on our dump), while, clearly, the correct entity should have been \emph{Beltran Levya}. We believe that this type of an error can be handled by performing joint EL and coreference resolution, which is a promising future research area for \liel.  

Contextual information can also hurt system performance \eg\ from the text, 
``\emph{.. dijo Alex S{\'a}nchez , analista..}'', \liel\ predicts
the Wikipedia title \texttt{Alex S{\'a}nchez (outfielder)} for the mention
\emph{Alex S{\'a}nchez} since the document talks about sports and
player names. The query mention was actually referring to a journalist, not in the KB, and hence a $NIL$. Handling sparse entities, similar to this, are also an important future direction.

\section{Related Work\label{sec:related-work} }

Entity linking has been introduced and actively developed under the
NIST-organized Text Analysis Conference, specifically the Knowledge
Base Population track. The top performing English EL system in the
TAC evaluation has been the MS\_MLI system \cite{sil13TAC},
which has obtained the top score in TAC evaluation in the past 4 years
(2011 through 2014): the system links all mentions in a document simultaneously,
with the constraint that their resolved links should be globally consistent on
the category level as much as possible.
Since global disambiguation can be expensive, \cite{2008-cikm-milne-witten-wikifier} uses the set of unambiguous
mentions in the text surrounding a mention to define the mention's
context, and uses the Normalized Google Distance \cite{2007-google-distance} to compute the similarity
between this context and the candidate Wikipedia entry. The UIUC system,
\cite{emnlp-chengroth-2013}, another state-of-the-art EL system, which
is an extension of \cite{2011-acl-illinois-wikifier}, adds relational
inference for wikification. \jnerd\ \cite{cikm-joint-nerel} is
a powerful joint entity extraction and linking system. However, by
construction their model is not language-independent due to the heavy
reliance on type systems of structured knowledge-bases like Freebase. It also
makes use of lexical features from Wikipedia as their model performs joint entity extraction
and disambiguation.
Some of the other systems which use a graph based algorithm such as
partitioning are LCC, NYU \cite{ji2014overview} and
HITS \cite{fahrni2013hits} which obtained competitive score in the
TAC evaluations. Among all these systems, only the HITS system has
ventured beyond English and has obtained the top score in Spanish
EL evaluation at TAC 2013. It is the only multilingual EL system in
the literature which performs reliably well across a series of languages
and benchmark datasets. Recently, \cite{wang2015language} show a new domain and language-independent EL system but they make use of translation tables for non-English (Chinese) EL; thereby not making the system
entirely language-independent. Empirically their performance comes close to System 1 which \liel\ outperforms.  The BASIS system \cite{merhav2013basis},
is the state-of-the-art for Chinese EL as it obtained the top score in
TAC 2013. The FUJITSU system \cite{fujitsu2013tac} obtained similar
scores. It is worth noting that these systems, unlike \liel, are heavily language
dependent, \eg\ performing lexicon specific information extraction,
using inter-language links to map between the languages or training
using labeled Chinese data. 

In more specialized domains, Dai \etal\ \shortcite{dai2011entity} employed a Markov
logic network for building an EL system with good results in a bio-medical
domain; it would be interesting to find out how their techniques might
extended to other languages/corpora. Phan \etal\ \shortcite{phan2008learning} utilize
topic models derived from Wikipedia to help classify short text segment,
while Guo \etal\ \shortcite{2013-naacl-twitter-nerd} investigate
methods for disambiguating entities in tweets. Neither of these methods
do show how to transfer the EL system developed for short texts to
different languages, if at all.

The large majority of entity linking research outside of TAC involves
a closely related task - wikification \cite{2006-eacl-bunescu-wikipedia-ne-disambig,2007-emnlp-conll-cucerzan-wikipedia-NE-disambig,2011-acl-illinois-wikifier,2013-naacl-twitter-nerd},
and has been mainly performed on English datasets, for obvious reasons
(data, tools availability). These systems usually achieve high accuracy
on the language they are trained on. Multilingual studies, e.g. \cite{mcnamee2011cross},
use a large number of pipelines and complex statistical machine translation
tools to first translate the original document contexts into English
equivalents and transform the cross-lingual EL task into a monolingual
EL one. The performance of the entity linking system is highly dependent
on the existence and potential of the statistical machine translation system in
the given pair of languages.

\section{Conclusion}

\label{sec:conclusion} In this paper we discussed a new strategy
for multilingual entity linking that, once trained on one language
source with accompanying knowledge base, performs without adaptation in multiple target languages. 
Our proposed system, \liel\, is trained on the English Wikipedia corpus, after building its own knowledge-base by exploiting the rich information present in Wikipedia. One of the
main characteristics of the system is that it makes effective use
of features that are built exclusively around computing similarity
between the text/context of the mention and the document text of the
candidate entity, allowing it to transcend language and perform inference
on a completely new language or domain, without change or adaptation.

The system displays a robust and strong empirical evidence by not only outperforming
all state-of-the-art English EL systems, but also achieving very good
performance on multiple Spanish and Chinese entity linking benchmark
datasets, and it does so without the need to switch,
retrain, or even translate, a major differentiating factor from the existing multi-lingual EL systems out there.

\section*{Acknowledgments}
We would like to thank the anonymous reviewers for their suggestions. We also thank Salim Roukos, Georgiana Dinu and Vittorio Castelli for their helpful comments. This work was funded under DARPA HR0011-12-C-0015 (BOLT). The views and findings in this paper are those of the authors and are not endorsed by  DARPA.

\bibliography{sil}
\bibliographystyle{acl2016}

\end{document}